\documentclass[sigconf,natbib=true,anonymous=false]{acmart}

\usepackage{algorithm}
\usepackage{algorithmic}
\usepackage{dcolumn,booktabs}
\usepackage{multirow}
\usepackage[flushleft]{threeparttable}
\usepackage{subfigure}
\usepackage{amsmath}
\usepackage{amsfonts}
\usepackage{bm}
\usepackage{graphicx}
\usepackage{amsthm}
\usepackage{enumitem}

\AtBeginDocument{%
  \providecommand\BibTeX{{%
    \normalfont B\kern-0.5em{\scshape i\kern-0.25em b}\kern-0.8em\TeX}}}

\copyrightyear{2023}
\acmYear{2023}
\setcopyright{acmlicensed}\acmConference[KDD '23]{Proceedings of the 29th ACM SIGKDD Conference on Knowledge Discovery and Data Mining}{August 6--10, 2023}{Long Beach, CA, USA}
\acmBooktitle{Proceedings of the 29th ACM SIGKDD Conference on Knowledge Discovery and Data Mining (KDD '23), August 6--10, 2023, Long Beach, CA, USA}
\acmPrice{15.00}
\acmDOI{10.1145/3580305.3599820}
\acmISBN{979-8-4007-0103-0/23/08}

\begin{document}

\title{Explicit Feature Interaction-aware Uplift Network for Online Marketing}

\author{Dugang Liu}
\authornote{This work was done during his internship at FiT, Tencent.}
\affiliation{%
  \institution{Guangdong Laboratory of Artificial Intelligence and Digital Economy (SZ), Shenzhen University}
  \city{Shenzhen}
  \country{China}}
\email{dugang.ldg@gmail.com}

\author{Xing Tang}
\authornote{Co-corresponding authors}
\affiliation{
  \institution{FiT, Tencent}
  \city{Shenzhen}
  \country{China}
}
\email{shawntang@tencent.com}

\author{Han Gao}
\affiliation{
  \institution{FiT, Tencent}
  \city{Shenzhen}
  \country{China}
}
\email{hansologao@tencent.com}

\author{Fuyuan Lyu}
\affiliation{
  \institution{McGill University}
  \city{Montreal}
  \country{Canada}
}
\email{fuyuan.lyu@mail.mcgill.ca}

\author{Xiuqiang He}
\authornotemark[2]
\affiliation{
 \institution{FiT, Tencent}
  \city{Shenzhen}
  \country{China}
}
\email{xiuqianghe@tencent.com}



\renewcommand{\shortauthors}{Liu, et al.}

\begin{abstract}
  As a key component in online marketing, uplift modeling aims to accurately capture the degree to which different treatments motivate different users, such as coupons or discounts, also known as the estimation of individual treatment effect (ITE). In an actual business scenario, the options for treatment may be numerous and complex, and there may be correlations between different treatments. In addition, each marketing instance may also have rich user and contextual features. However, existing methods still fall short in both fully exploiting treatment information and mining features that are sensitive to a particular treatment. In this paper, we propose an explicit feature interaction-aware uplift network (EFIN) to address these two problems. Our EFIN includes four customized modules: 1) a feature encoding module encodes not only the user and contextual features, but also the treatment features; 2) a self-interaction module aims to accurately model the user's natural response with all but the treatment features; 3) a treatment-aware interaction module accurately models the degree to which a particular treatment motivates a user through interactions between the treatment features and other features, i.e., ITE; and 4) an intervention constraint module is used to balance the ITE distribution of users between the control and treatment groups so that the model would still achieve a accurate uplift ranking on data collected from a non-random intervention marketing scenario. We conduct extensive experiments on two public datasets and one product dataset to verify the effectiveness of our EFIN. In addition, our EFIN has been deployed in a credit card bill payment scenario of a large online financial platform with a significant improvement.
\end{abstract}





\maketitle

\section{Introduction}\label{sec:intro}
To increase the user engagement and platform revenue, providing some specific incentives to the users, such as coupons~\cite{zhao2019unified}, discounts~\cite{lin2017monetary}, and bonuses~\cite{ai2022lbcf}, is an important strategy in online marketing~\cite{reutterer2006dynamic}.
Since these incentives usually have a cost and different users have different responses to these incentives, such as some users do not consume without a coupon and some users will consume anyway, how to accurately identify the corresponding sensitive user groups for each incentive is critical to maximize marketing benefits~\cite{he2022causal,xu2022learning}.
To achieve this goal, we need to accurately capture the difference between users' responses to various incentives compared to those without incentives.
Unlike traditional supervised learning, this involves a typical causal inference problem, because in a practical scenario, we can usually only observe one type of the user responses, which may be for a certain incentive (i.e., treatment group) or for no incentive (i.e., control group).
Therefore, the change in the user's response caused by different incentives (or treatments) that we want to obtain can be regarded as the estimation of the individual treatment effect (ITE)~\cite{zhang2021unified}, also known as the uplift.
To solve the above estimation problem, in recent years, uplift modeling has been proposed and its effectiveness has been verified~\cite{devriendt2020learning,betlei2021uplift,chen2022imbalance}.

\begin{figure*}[htbp]
\centering
\includegraphics[width=1.\textwidth]{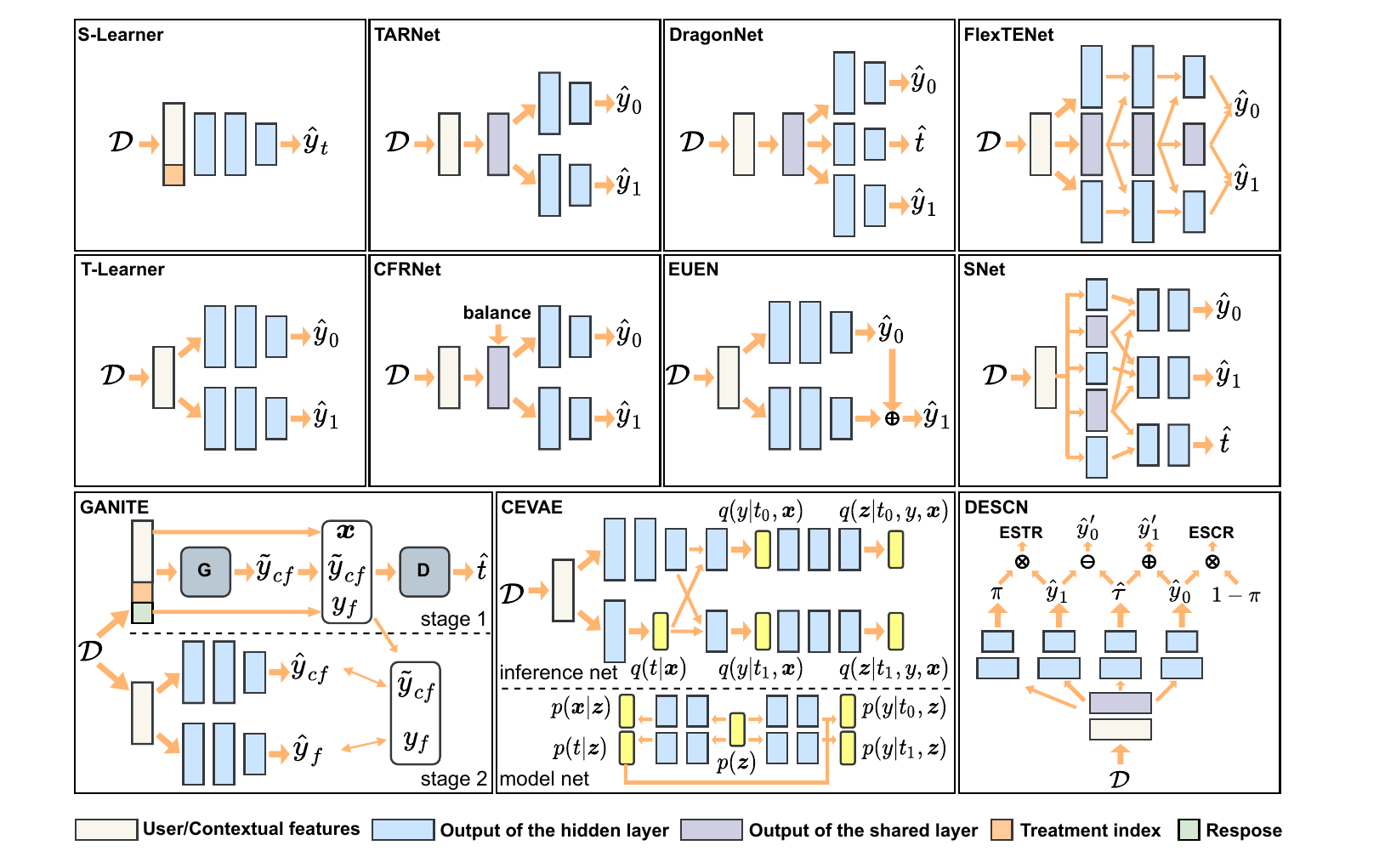}
\Description[<Figure 1. Fully described in the text.>]{<In Figure 1, we show the model architectures of some representative methods in neural network-based uplift modeling.>}
\caption{
Architecture diagram of some representative methods in neural network-based uplift modeling, where $\mathcal{D}$ denotes a training set, $\hat{y}_0$ and $\hat{y}_1$ denote the predicted response for the control and treatment groups, respectively, and $\hat{t}$ denotes the predicted label of the treatment. In GANITE, G, D, $\widetilde{y}_{cf}$, and $\hat{y}_{cf}$ are the generator, the discriminator, the generated counterfactual response and the corresponding predicted response, respectively. In CEVAE, $p()$ and $q()$ denote different distributions in the inference network and the model network, respectively. In DESCN, $\hat{\tau}$, $\pi$, $\hat{y}'_0$, and $\hat{y}'_1$ denote the predicted ITE, the probability that an instance belongs to the treatment group, and the predicted cross-control and cross-treatment responses, respectively.
}
\label{fig:1}
\end{figure*}

The existing uplift modeling methods mainly includes three research lines according to the design ideas:
1) \textit{Meta-learner based.} The basic idea of this line is to use the existing prediction methods to build the estimator for the users' responses, which may be global (i.e., S-Learner) or divided by the treatment and control groups (i.e., T-Learner)~\cite{kunzel2019metalearners}. Based on this, different two-step learners can be designed by introducing various additional operations, such as X-Learner~\cite{kunzel2019metalearners}, R-Learner~\cite{nie2021quasi}, and DR-Learner~\cite{bang2005doubly}, etc.
2) \textit{Tree based.} The basic idea of this line is to use a tree structure to gradually divide the entire user population into the sub-populations that are sensitive to each treatment. The key step is to directly model the uplift using different splitting criteria, such as based on various distribution divergences~\cite{radcliffe2011real} and the expected responses~\cite{zhao2017uplift,saito2020cost}. In addition, causal forest~\cite{athey2016recursive} obtained by integrating multiple trees is another representative method on this line, and several variants have been proposed~\cite{wan2022gcf,ai2022lbcf}.
3) \textit{Neural network based.} The basic idea of this line is to take advantage of neural networks to design more complex and flexible estimators for the user's response~\cite{louizos2017causal,yoon2018ganite,ke2021addressing,zhong2022descn}, and most of them can be seen as improvements of the T-learner~\cite{shalit2017estimating,shi2019adapting,curth2021inductive,curth2021nonparametric}. 
In this paper, we focus on neural network-based line because it can be better adapted to the goal of feature interaction modeling introduced in this paper due to the flexibility of neural networks.
Also, since various neural network models are commonly employed in commercial systems, research on this line can be more easily integrated than other lines.
We present the architectures of some representative methods in neural network-based uplift modeling in Figure~\ref{fig:1}.

Although existing uplift modeling methods have shown promising results, most of them still fall short in both fully exploiting treatment information and mining features that are sensitive to a particular treatment.
In an online marketing, the treatment usually has many features that describe it in detail in addition to the index ID.
For example, a coupon may include a specific amount and a minimum spending amount to be reached.
This also means that different treatments may be related, such as having similar amounts or minimum spending to be achieved.
Intuitively, this information is beneficial for obtaining a accurate uplift, e.g., the correlation between the treatments can prompt the model to discover that a user's response to a coupon worth 1000 should be more similar to a coupon worth 900 than to a coupon worth 100.
However, as shown in Figure~\ref{fig:1}, we can find that almost all related methods do not explicitly utilize treatment features, which may be detrimental to the uplift estimation.
We refer to this challenge as \textbf{underutilization of treatment features}.
Furthermore, the above challenges will also prevent most related methods from accurately capturing the sensitive features associated with each treatment, due to the lack of modeling of the interactions between treatment features and the rest.
We refer to this challenge as \textbf{underutilization of feature interactions}.
Note that explicitly modeling the treatment features may also make the model compatible with a variety of marketing scenarios, where treatment options may be binary, multi-valued, or continuous, without significantly increasing the size of the model.

To address the above two challenges, in this paper, we propose an explicit feature interaction-aware uplift network (EFIN).
Specifically, our EFIN includes four modules:
1) a feature encoder module aims to encode a marketing instance containing the user features, the contextual features, and the treatment features;
2) a self-interaction module is responsible for the responses of the users in the control group. It uses a self-attention network to model the interactions between all the features except the treatment features to capture a subset of features associated with the natural responses (i.e., not receiving any the treatment);
3) a treatment-aware interaction module is responsible for the responses of users in the treatment group. It uses a treatment-aware attention network to model the interaction between the treatment features and other features to identify subsets of features that are sensitive to different treatments, and to accurately capture a user's changes in response to different treatments;
and 4) an intervention constraint module is used to balance the ITE distribution of users between the treatment and control groups so that our EFIN could be more robust in different scenarios. This module is necessary since the treatment assignment is usually non-random in a real marketing scenario and will result in differences in user distribution between control and treatment groups.
Finally, we conduct extensive offline and online evaluations and the results validate the effectiveness of our EFIN.

\section{Related Work}\label{sec:related}
In this section, we briefly review some relevant works on two research topics, including uplift modeling and feature interaction.

\subsection{Uplift Modeling}
Uplift modeling aims to identify the corresponding sensitive population for each specific treatment by accurately estimating ITE.
The existing uplift modeling methods mainly includes three research lines: 
1) a meta-learner-based method focuses on using existing prediction methods to learn a one-step learner~\cite{kunzel2019metalearners} or a two-step learner~\cite{nie2021quasi,bang2005doubly} for the user's response, where the treatment information is usually integrated as one-dimensional discrete features or as a prior for switching prediction branches.
2) a tree-based method employs a specific tree or a forest structure with splitting criterion of different metrics to gradually divide the sensitive subpopulations corresponding to each treatment from the entire population~\cite{radcliffe2011real,zhao2017uplift,athey2016recursive}, where the treatment information is included in the calculation of the splitting process;
and 3) a neural network-based method combines the advantages of neural networks to introduce some more complex and flexible architectures to model the response process to the treatment, which can learn a more accurate estimator for the users' responses or the uplifts.
Furthermore, there are only a few works that address uplift modeling by linking it to the well-established problems in other fields, such as the knapsack problem~\cite{goldenberg2020free,albert2021commerce}.
Our EFIN follows a neural network-based line, but differs significantly from existing related works, especially in the explicit utilization of the treatment feature and the modeling of its interactions with other features.

\subsection{Feature Interaction}
Feature interactions are designed to model combinations between different features and have been shown to significantly improve the performance of a response model~\cite{Optinter,lyu2023optimizing}.
Existing feature interaction methods can be mainly divided into three categories, including second-order interactions, higher-order interactions, and structural interactions.
In second-order interactions, the inner product between the embedding representations of two features is usually considered, and factorization machines and their variants are representative methods~\cite{rendle2011fast,juan2016field}.
Modeling of higher-order interactions relies on neural networks, and many architectures have been proposed to enhance model performance, interpretability, and efficient fusion of lower- and higher-order interactions~\cite{guo2017deepfm,wang2021dcn}.
In addition, based on the graph structure, some methods aim to exploit the additional structural information to further improve higher-order interactions~\cite{li2019fi,liu2022user}.
Although feature interaction has achieved success on many tasks, research on its application in uplift modeling is still lacking.
Our EFIN aims to bridge the gap in this research direction.

\section{Preliminaries}\label{sec:preliminaries}
Let $\{z_i\}_{i=1}^{n}=\{(\bm{x}_i,\bm{t}_i,y_i)\}_{i=1}^{n} \in \mathcal{X} \times  \mathcal{T} \times \mathcal{Y}$ denote a marketing instance, where $\bm{x}_i=\left[x_{i0}, x_{i1}, \dots,x_{i{d_{x-1}}},x_{i{d_x}}\right]$ is the $d_x$-dimensional user features and contextual features included in the $i$-th instance, $\bm{t}_i=\left[t_{i0},t_{i1},\dots,t_{i{d_{t-1}}},t_{i{d_t}}\right]$ is the $d_t$-dimensional treatment features included, $n$ is the number of the training instances, and $y_i\in\{0,1\}$ is the response label for the $i$-th instance.
Without loss of generality, we assume that the first treatment feature of each instance denotes the index ID of the treatment, and the total number of treatments is $K$, i.e., $t_{i0}\in\{0,1,\dots,K\}$.

Following the Neyman-Rubin potential outcome framework~\cite{rubin2005causal}, let $y_i(k)$ and $y_i(0)$ denote the potential outcome when the user in the $i$-th instance gets a particular treatment $k\in\{1,\dots,K\}$ or is not treated, respectively.
The probability of each treatment being assigned can be denoted as $\pi_k(\bm{x}_i)=P(t_{i0}=k|\bm{x}_i)$, also known as a propensity score.
Since we can usually only observe $y_i(k)$ or $y_i(0)$, but not both, i.e., $y_i=y_i(k)$ or $y_i=y_i(0)$, there is no true uplift result $y_i(k)-y_i(0)$ for each instance, which is a key reason uplift modeling differs from traditional supervised learning.
Therefore, uplift modeling aims to accurately estimate the expected individual treatment effect $\tau_k(\bm{x}_i)$ for each instance.
Specifically, following standard assumptions~\cite{zhong2022descn}, this estimate can be expressed as,
\begin{equation}\label{equ:1}
\begin{aligned}
	\tau_k(x_i)&=\mathbb{E}(y_i(k)-y_i(0)|x_i),\\
	&=\mathbb{E}(y_i(k)|t_{i0}=k,x_i)-\mathbb{E}(y_i(0)|t_{i0}=0,x_i).
\end{aligned}
\end{equation}
After obtaining all the estimated individual treatment effects $\tau_k(x_i)$, we can rank them and make a rational treatment assignment.

\section{The Proposed Method}\label{sec:efin}

\begin{figure*}[htbp]
\centering
\includegraphics[width=0.95\textwidth]{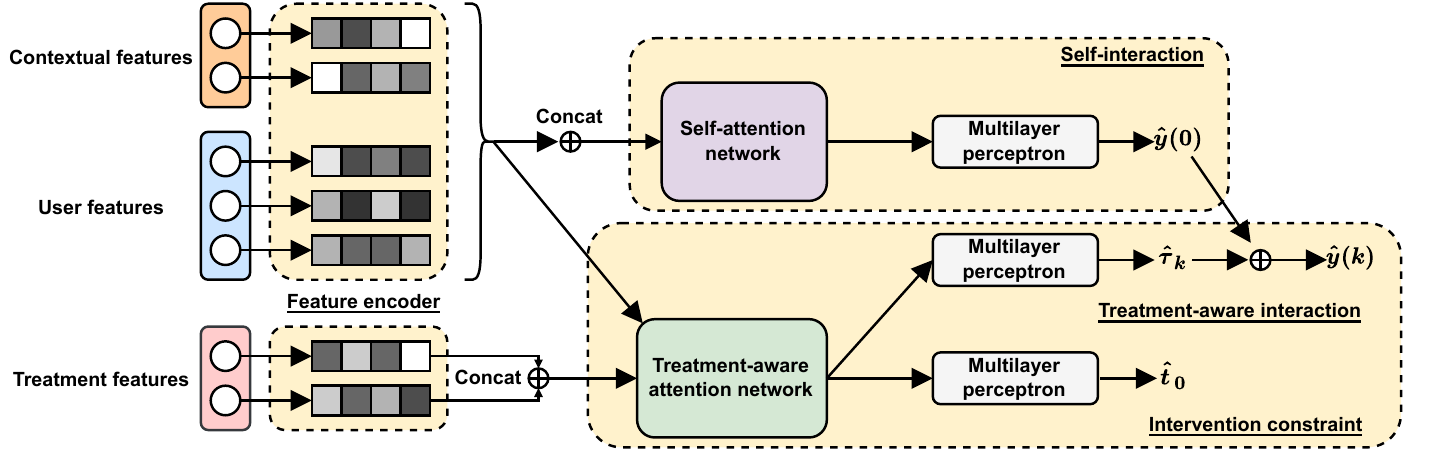}
\Description[<Figure 2. Fully described in the text.>]{<A full description of Figure 2 can be found in Section 4.1 of the text.>}
\caption{The architecture of the explicit feature interaction-aware uplift network (EFIN)}
\label{fig:2}
\end{figure*}

\subsection{Architecture}\label{sec:efin:architecture}
As mentioned in Section~\ref{sec:intro}, most of the existing methods generally suffer from two challenges of underutilization of treatment features and feature interactions.
To address the above two challenges, in this paper, we propose an explicit feature interaction-aware uplift network (EFIN) and illustrate the architecture of our EFIN in Figure~\ref{fig:2}.
Given a current marketing instance $z_i=(\bm{x}_i,\bm{t}_i,y_i)$, the feature encoder module will encode non-treatment features $\bm{x}_i$ and treatment features $\bm{t}_i$ separately to obtain their respective embedding representations, i.e., $\mathbf{e}^x$ and $\mathbf{e}^t$.
The embedded representation $\mathbf{e}^x$ will be fed into a self-interaction module with a self-attention network and multiple multilayer perceptrons, which computes the natural response of the instance when it is not treated, i.e., $\hat{y}_i(0)$.
The embedded representations $\mathbf{e}^x$ and $\mathbf{e}^t$ will be fed into a treatment-aware interaction module to compute the ITE this instance has for a particular treatment, i.e., $\hat{\tau}_k(\bm{x}_i)$, where a treatment-aware attention network will model the interaction of $\mathbf{e}^x$ and $\mathbf{e}^t$.
In addition, the estimated ITE will be combined with the previously predicted natural response to generate the response of this instance to a particular treatment, i.e., $\hat{y}_i(k)=\hat{y}_i(0)+\hat{\tau}_k(\bm{x}_i)$.
The input of an intervention constraint module is the embedded representation $\mathbf{e}^{xt}$ containing interaction information obtained after going through the treatment-aware attention network, and the goal is to predict the group to which this instance belongs, i.e., $\hat{t}_{i0}$.
The final optimization objective function of our EFIN can be expressed as follows,
\begin{equation}\label{equ:2}
\mathop{\min}\limits_{\theta} \mathcal{L}_{EFIN} = \mathcal{L}_{S} + \mathcal{L}_{T} + \mathcal{L}_{C} + \lambda \lVert\theta\rVert,
\end{equation}
where $\mathcal{L}_{S}$, $\mathcal{L}_{T}$, and $\mathcal{L}_{C}$ denote the losses for the self-interaction module, treatment-aware interaction module, and intervention constraint module, respectively, and $\lambda$ and $\lVert\theta\rVert$ are the tradeoff parameter and the regularization terms.

\subsection{Training}\label{sec:efin:training}
In this subsection, we describe each module in detail based on the training process.

\subsubsection{The Feature Encoder Module}\label{sec:efin:training:feature}
Given a current marketing instance $z_i=(\bm{x}_i,\bm{t}_i,y_i)$, unlike most existing works, we encode not only non-treatment features $\bm{x}_i$ but also treatment features $\bm{t}_i$ in this module.
Taking the treatment feature as an example, for each continuous feature $t_{ij}$ in the treatment features, we equip it with a shared fully-connected network for encoding.
For the remaining features in the treatment features, i.e., the sparse features, we initialize an embedding table for each feature and obtain the embedding representation corresponding to the specific feature value through the \textit{lookup} operation.
This encoding process can be expressed as,
\begin{equation}\label{equ:3}
\mathbf{e}^t_{ij}=\begin{cases}\mathbf{W}_j*t_{ij} + \mathbf{b}_j, & \text{ $t_{ij}$ is a continuous feature,} \\
lookup(\mathbf{E}_j,e_{ij}), & \text{ $t_{ij}$ is a sparse feature,}
\end{cases}
\end{equation}
where $\mathbf{W}_j$ is a weight matrix, $\mathbf{b}_j$ is a bias vector, and $\mathbf{E}_j$ is an embedding table.
Intuitively, continuous features in treatment features are more likely to reflect the correlation between different treatments, such as similar amounts and minimum consumption to be satisfied, so the different encoding of continuous features in Eq.\eqref{equ:3} aims to preserve this property.
For non-treatment features, we adopt a similar encoding process. 
Finally, we can obtain the corresponding embedding representation, i.e., $\mathbf{e}^x_i=\{\mathbf{e}^x_{i0}, \mathbf{e}^x_{i1},\dots,\mathbf{e}^x_{id_{x}}\}$ and $\mathbf{e}^t_i=\{\mathbf{e}^t_{i0}, \mathbf{e}^t_{i1},\dots,\mathbf{e}^t_{id_{t}}\}$.

\subsubsection{The Self-interaction Module}\label{sec:efin:training:self}
In this module, we use the embedding representation $\mathbf{e}^x_i$ to model the natural response of each user in the control group, where information about the treatment is isolated to capture user-sensitive features in the natural situation.
We use a self-attention network for self-interaction to better predict natural responses.
Specifically, we have,
\begin{equation}\label{equ:4}
    Q = K = V = \left(\mathbf{e}^x_{i0};\mathbf{e}^x_{i1};\dots;\mathbf{e}^x_{i2};\dots;\mathbf{e}^x_{id_{x}}\right),\\
\end{equation}
\begin{equation}\label{equ:5}
\mathbf{\overline{e}}^{x}_i = softmax(\frac{QK^T}{\sqrt{K_d}})V,
\end{equation}
where $K_d$ is the dimension of the output embedding, and $\mathbf{\overline{e}}^{x}_i=\{\mathbf{\overline{e}}^{x}_{i0}, \mathbf{\overline{e}}^{x}_{i1},\dots,\mathbf{\overline{e}}^{x}_{id_{x}}\}$. 
Next, we use a multilayer perceptron to predict natural responses,
\begin{equation}\label{equ:6}
\hat{y}_i(0)=\mathbf{W}_s*concat(\mathbf{\overline{e}}^{x}_i) + \mathbf{b}_s,
\end{equation}
where $\mathbf{W}_s$ is a weight matrix and $\mathbf{b}_s$ is a bias vector. 
Based on the control group instances contained in the training set, the optimization objective of the self-interaction module is a supervised loss for natural responses,
\begin{equation}\label{equ:7}
\mathcal{L}_{S} = \mathcal{L}(\hat{y}_i(0),y_i(0)).
\end{equation}

\subsubsection{The Treatment-aware Interaction Module}\label{sec:efin:training:treatment}
In this module, we aim to use the embedding representations $\mathbf{e}^x_i$ and $\mathbf{e}^t_i$ to learn the responses of users in different treatment groups and to identify the corresponding sensitive features, where the treatment information will be used as inducements to achieve this goal. 
Note that this is different from the self-interaction module.
Specifically, we first use a treatment-aware attention network to model the interaction between treatment features and non-treatment features, and use attention weights to describe the sensitivity of non-treatment features to a particular treatment,
\begin{align}\label{equ:8}
	\alpha^i_j=\operatorname{Softmax}(\mathbf{W}_{t0}^\top&\operatorname{Relu}(\mathbf{W}_{t1} \mathbf{e}^t_i + \mathbf{W}_{t2} \mathbf{e}^x_{ij} + \mathbf{b}_{t2})),\\
	\mathbf{e}^{xt}_i&=\sum_{j=1}^{d_x} \alpha^i_j \mathbf{e}^x_{ij},
\end{align}
where $\mathbf{W}_{t0}$, $\mathbf{W}_{t1}$ and $\mathbf{W}_{t2}$ are weight matrices and $\mathbf{b}_{t2}$ is a bias vector.
Based on this embedded representation combined with interaction information, we then estimate the ITE of users in different treatment groups,
\begin{equation}\label{equ:9}
\hat{\tau}_k({\mathbf{x}_i})=\mathbf{W}_{t3}*\mathbf{e}^{xt}_i + \mathbf{b}_{s3},
\end{equation}
where $\mathbf{W}_{t3}$ is a weight matrix and $\mathbf{b}_{t3}$ is a bias vector.
By combining the estimated ITE with the natural responses predicted by Eq.\eqref{equ:6}, we can obtain the responses of users in different treatment groups, and construct the supervised loss of this module with instances of different treatment groups in the training set,
\begin{equation}\label{equ:10}
\hat{y}_i(k)=\hat{y}_i(0)+\hat{\tau}_k(\bm{x}_i).
\end{equation}
\begin{equation}\label{equ:11}
\mathcal{L}_{T} = \mathcal{L}(\hat{y}_i(k),y_i(k)).
\end{equation}

\subsubsection{The Intervention Constraint Module}\label{sec:efin:training:intervention}
\begin{figure}[htbp]
\centering
\includegraphics[width=0.77\columnwidth]{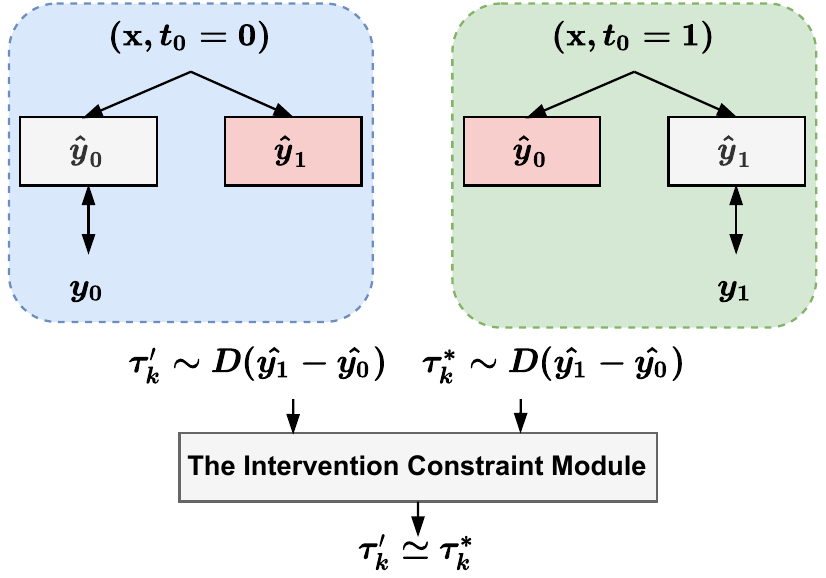}
\Description[<Figure 3. Fully described in the text.>]{<A full description of Figure 3 can be found in Section 4.2.4 of the text.>}
\caption{Illustration of the idea behind the intervention constraint module.}
\label{fig:3}
\end{figure}

Since in an online marketing scenario, the assignment of different treatments is usually not random, and this means that the collected training set usually has a significant distribution difference between the control and treatment groups.
As shown in Figure~\ref{fig:3}, since only one type of response in each group is available for supervised training, differences in distribution between groups will exacerbate significant differences in estimated ITE between groups, such as $\tau'_k$ and $\tau^*_k$.
Therefore, ignoring this difference may increase the difficulty of ITE estimation and impair accuracy.
To alleviate this problem, we propose a simple but effective intervention constraint module.
The idea behind this module is to increase the difficulty of guessing the corresponding group from the ITE distribution of different groups, that is, to achieve a trade-off between the two through mutual interference.
Previous studies have shown that similarities in ITE distribution across groups are beneficial for uplift modeling~\cite{curth2021inductive}.
Specifically, we use the embedding representation $\mathbf{e}^{xt}_i$, which is closely related to the uplift, to make predictions about which group this instance belongs to.
We then train it with an inverse group label to generate perturbations as described above.
This process can be expressed as,
\begin{equation}\label{equ:12}
\hat{t}_{i0}=\mathbf{W}_{c}*\mathbf{e}^{xt}_i + \mathbf{b}_{c}.
\end{equation}
\begin{equation}\label{equ:13}
\mathcal{L}_{C} = \mathcal{L}(\hat{t}_{i0},\overline{t}_{i0}).
\end{equation}
Note that in the case of binary treatment, $\overline{t}_{i0}$ can directly take the opposite label.
In the case of multi-valued treatment, a 0-1 mask vector needs to be generated from the original labels, and then the labels are negated.

\subsubsection{The Uplift Prediction}\label{sec:efin:training:uplift}
After our EFIN training is complete, in the inference phase, we only need to use the treatment-aware interaction module to directly compute ITE, and then perform ranking and decision-making.

\section{Empirical Evaluations}
In this section, we conduct experiments with the aim of answering the following three key questions. 
\begin{itemize}[leftmargin=*]
    \item RQ1: How does our EFIN perform compared to the baselines?
    \item RQ2: What is the role of each module in our EFIN?
    \item RQ3: How effective is our EFIN in an online deployment?
\end{itemize}

\subsection{Experimental Setup}

\subsubsection{Datasets.}
Following the settings of a previous work~\cite{ke2021addressing}, we conduct experiments on two public datasets including CRITEO-UPLIFT~\cite{diemert2021large} and EC-LIFT~\cite{ke2021addressing}.
CRITEO-UPLIFT is a dataset open sourced by Criteo AI Labs for uplift modeling in a large-scale advertising scenario, which includes nearly 14 million instances, twelve continuous features, and binary treatments.
EC-LIFT is a dataset of uplift modeling for different brands in a large-scale advertising scene, which is open sourced by Alimama. 
This dataset contains billions of instances, twenty-five discrete features and nine multi-valued features, and binary treatments.
Due to the excessively large data scale, in order to facilitate training, we extracted about 40\% of the instances from the original EC-LIFT dataset as the experimental dataset.
The statistics of the two public datasets are shown in Table~\ref{tab:datasets}.
We randomly split the two dataset for training and testing with a ratio 8/2.
Note that since the modeling of user features and contextual features is consistent and does not require a special distinction, we use all the features in the above datasets in our experiments.
Furthermore, following the setting of previous work, we treat treatment as a binary feature.
To comprehensively evaluate our EFIN, we also include a product dataset collected from two weeks of online coupon marketing scenarios for credit card repayments.
This product dataset uses a total of more than 200 features, involves 2 million users and has 2 million instances, 90\% of which are used for the training set and the rest for the test set.
In particular, instead of binary treatments in the public datasets, seven treatment options are included in product dataset.
\begin{table}[htbp]
\caption{Statistics of the two public datasets.}
\centering
\scalebox{0.87}{
\begin{tabular}{l|l|l}
\specialrule{0.1em}{3pt}{3pt}
\textbf{Dataset} & \textbf{CRITEO-UPLIFT} & \textbf{EC-LIFT} \\
\specialrule{0.05em}{3pt}{3pt}
Size & 13,979,592 & 196,084,380 \\
Ratio of Treatment to Control & 5.67:1 & 3.11:1 \\
Average Visit Ratio & 4.70\% & 3.25\% \\
Relative Average Uplift & 27.07\% & 464.46\% \\
Average Uplift & 1.03\% & 3.56\% \\
Conversion Target & Visit & Visit \\
\specialrule{0.1em}{3pt}{3pt}
\end{tabular}}
\label{tab:datasets}
\end{table}

\subsubsection{Evaluation Metrics} 
We evaluate uplift ranking performance by four widely used metrics, i.e., uplift score at first $h$ percentile (LIFT@$h$), normalized area Under the uplift curve (AUUC), normalized area under the qini curve (QINI) and Weighted average uplift (WAU).
We report the results with $h$ set to 30.
We use a standard python package \textit{scikit-uplift}\footnote{https://www.uplift-modeling.com/en/latest/} to compute these metrics.

\subsubsection{Baselines} 
To evaluate the effectiveness of our EFIN, we select a set of the most representative methods in neural network-based uplift modeling, including S-Learner~\cite{kunzel2019metalearners}, T-Learner~\cite{kunzel2019metalearners}, TarNet~\cite{shalit2017estimating}, CFRNet~\cite{shalit2017estimating}, DragonNet~\cite{shi2019adapting}, GANITE~\cite{yoon2018ganite}, CEVAE~\cite{louizos2017causal}, SNet~\cite{curth2021nonparametric}, FlexTENet~\cite{curth2021inductive}, EUEN~\cite{ke2021addressing} and DESCN~\cite{zhong2022descn}.

\subsubsection{Implementation Details}
We implement all baselines and our EFIN in PyTorch 1.13\footnote{https://pytorch.org/}.
We use an AdamW optimizer\footnote{https://pytorch.org/docs/stable/generated/torch.optim.AdamW.html} and set the maximum number of iterations to 20.
To search for the best hyperparameters, we use QINI as a primary evaluation metric.
We also adopt an early stopping mechanism with a patience of 5 to avoid overfitting to the training set.
Furthermore, we use the hyper-parameter search library \textit{Optuna}\footnote{https://optuna.org/} to accelerate the tuning process.
The range of the values of the hyper-parameters are shown in Table~\ref{tab:search_ranges}.
\begin{table}[htbp]
\caption{Hyper-parameters and their values tuned in the experiments.}
\centering
\scalebox{0.9}{
\begin{tabular}{ccc}
\specialrule{0.1em}{3pt}{3pt}
\textbf{Name} & \textbf{Range} & \textbf{Functionality}\\
\specialrule{0.05em}{3pt}{3pt}
$rank$ & $\left\{2^{5},2^{6},2^{7}\right\}$ & Embedded dimension \\
$bs$ &$\left\{2^{8},2^{9},2^{10},2^{11}\right\}$ & Batch size \\
$lr$ &$\left\{1e^{-4},1e^{-3},1e^{-2},1e^{-1}\right\}$ & Learning rate \\
$\lambda$ &$\left\{1e^{-5},1e^{-4},1e^{-3},1e^{-2},1e^{-1}\right\}$ & Loss weighting \\
\specialrule{0.1em}{3pt}{3pt}\\
\end{tabular}}
\label{tab:search_ranges}
\end{table}

\subsection{RQ1: Performance Comparison}\label{sec:experiments:performance}
We report the comparison results on two public datasets in Table~\ref{tab:main_result}.
From the results in Table~\ref{tab:main_result}, we can have the following observations: 
1) T-learner significantly outperform S-learner, even some baselines using more complex network architectures. This may mean that in online marketing scenarios with numerous features, uplift modeling is more difficult than traditional ITE estimation, especially the sensitive features of users need to be more accurately identified. In particular, we can observe that on EC-LIFT with a large number of high-dimensional sparse features, most baselines no longer have an advantage over the T-learner.
2) By designing some more flexible or complex architectures as estimators of user responses, FlexTENet, SNet, EUEN and DESCN perform better than other baselines. But again, their advantage over EC-LIFT shrinks. This means that other architectural changes may not yield much gain without taking feature interactions into account.
3) Unlike other baselines, our EFIN consistently outperforms all baselines in most cases except slightly weaker than DESCN on WAU. Since we use QINI as the main metric in the hyperparameter search, there may be some fluctuations in other metrics, and we can find that our EFIN has a large improvement on QINI. Furthermore, our EFIN is also able to maintain the performance advantage on EC-LIFT, benefiting from the explicit modeling of treatment features and feature interactions.

\begin{table*}[htbp]
\caption{Results on two public datasets, where the best and second best results are marked in bold and underlined, respectively. Note that $^{*}$ indicates a significance level of $p\leq 0.05$ based on two-sample t-test between our method and the best baseline.}
\centering
\scalebox{1.1}{
\begin{tabular}{c|cccc|cccc}
\specialrule{0.1em}{3pt}{3pt}
 {Dataset} & \multicolumn{4}{c|}{CRITEO-UPLIFT} & \multicolumn{4}{c}{EC-LIFT} \\
 \specialrule{0.05em}{1pt}{1pt}
 Metrics & LIFT@30 & QINI & AUUC & WAU  & LIFT@30  & QINI  & AUUC & WAU  \\
\specialrule{0.05em}{1pt}{3pt}
S-Learner & 0.0328 & 0.0857 & 0.0332 & 0.0092 & 0.0080 & 0.0414 & 0.0073 & 0.0031 \\
T-Learner & 0.0425 & 0.1083 & 0.0430 & 0.0093 & 0.0086 & 0.0440 & 0.0079 & 0.0032 \\
\hline
TarNet & 0.0339 & 0.1027 & 0.0406 & 0.0087 & 0.0081 & 0.0422 & 0.0076 & 0.0031 \\
CFRNet & 0.0379 & 0.1052 & 0.0414 & 0.0101 & 0.0087 & 0.0422 & 0.0078 & 0.0031 \\
DragonNet & \underline{0.0464} & 0.1096 & 0.0437 & 0.0093 & \underline{0.0096} & \underline{0.0459} & \underline{0.0092} & 0.0033 \\
GANITE & 0.0447 & \underline{0.1170} & \underline{0.0468} & 0.0101 & 0.0080 & 0.0409 & 0.0068 & 0.0029 \\
CEVAE & 0.0365 & 0.0951 & 0.0375 & 0.0106 & 0.0077 & 0.0373 & 0.0068 & 0.0031 \\
FlexTENet & 0.0448 & 0.1108 & 0.0441 & 0.0093 & 0.0084 & 0.0435 & 0.0078 & 0.0031 \\
SNet & 0.0442 & 0.1112 & 0.0442 & 0.0083 & 0.0084 & 0.0441 & 0.0079 & 0.0032 \\
EUEN & 0.0425 & 0.1153 & 0.0457 & 0.0108 & 0.0090 & 0.0446 & 0.0084 & 0.0033 \\
DESCN & 0.0456 & 0.1129 & 0.0455 & $\textbf{0.0131}^{*}$ & 0.0082 & 0.0435 & 0.0075 & \underline{0.0034} \\
\specialrule{0.05em}{1pt}{1pt}
EFIN & $\textbf{0.0468}^{*}$ & $\textbf{0.1285}^{*}$ & $\textbf{0.0514}^{*}$ & \underline{0.0122} & $\textbf{0.0100}^{*}$ & $\textbf{0.0468}^{*}$ & $\textbf{0.0097}^{*}$ & \textbf{0.0034} \\
\specialrule{0.1em}{1pt}{1pt}
\end{tabular}}
\label{tab:main_result}
\end{table*}

Next, we report the comparison results on the product dataset in Table~\ref{tab:main_result_1}.
Since most of the baselines are usually only applied to binary treatment scenarios, to evaluate them on the product dataset with multi-valued treatments, we first extend them reasonably, such as the network architecture changing from two-head to multi-head.
Note that since the distribution estimation in CEVAE is difficult to directly extend to the multi-head architecture, we do not report its results on the product dataset.
After the expansion is complete, we retrain all the methods using the same search range as in Table~\ref{tab:search_ranges}.
Note that when evaluating, we need to treat multi-valued treatments as multiple binary treatments to obtain individual metrics for each treatment, and finally report the averaged results.
From the results in Table~\ref{tab:main_result_1}, we have the following observations:
1) meta-learner-based methods (S-Learner and T-Learner) are still relatively stable and have suboptimal results in multi-valued treatments scenarios.
2) the baselines considering a shared architecture suffer from a performance bottleneck, where the shared part may cause learning shocks due to too many and significantly different treatment groups.
3) similarly, since our EFIN exploits treatment features and feature interactions explicitly, on the product dataset it still retains the ability to mine for each user its sensitive features associated with different treatments.
Combined with the results on two public datasets and one product dataset, this both validates the effectiveness of our EFIN, especially considering treatment features and feature interactions explicitly in the uplift modeling.

\begin{table}[htbp]
\caption{Results on a product dataset, where the best and second best results are marked in bold and underlined, respectively. Note that $^{*}$ indicates a significance level of $p\leq 0.05$ based on two-sample t-test between our method and the best baseline.}
\centering
\scalebox{1.1}{
\begin{tabular}{c|cc}
\specialrule{0.1em}{3pt}{3pt}
 {Dataset} & \multicolumn{2}{c}{Product}\\
\specialrule{0.05em}{1pt}{1pt}
 Metrics & Average QINI & Average AUUC \\
\specialrule{0.05em}{3pt}{3pt}
S-Learner & 0.0155 & $\textbf{0.0094}^{*}$\\
T-Learner &  \underline{0.0158} & 0.0034\\
\hline
TarNet & 0.0118 & 0.0001 \\
CFRNet & 0.0110 & 0.0066 \\
DragonNet & 0.0136 & 0.0004\\
GANITE & 0.0101 & 0.0047 \\
CEVAE & - & - \\
FlexTENet & 0.0143 & 0.0076 \\
SNet & 0.0122 & 0.0064 \\
EUEN & 0.0088 & 0.0003 \\
DESCN & 0.0128 & 0.0003 \\
\specialrule{0.05em}{1pt}{1pt}
EFIN & $\textbf{0.0172}^{*}$ & \underline{0.0085} \\
\specialrule{0.1em}{3pt}{3pt}
\end{tabular}}
\label{tab:main_result_1}
\end{table}

\subsection{RQ2: Ablation Study of EFIN}\label{sec:experiments:ablation}
Moreover, we conduct ablation studies of our EFIN to analyze the role played by each proposed module.
We sequentially removed the three core modules individually, i.e., the self-interaction module, the treatment-aware interaction module, and the intervention constraint module.
The results are shown in Table~\ref{tab:ablation}.
From the results in Table~\ref{tab:ablation}, we can find that removing any module will bring performance degradation.
This verifies the validity of each module design in our EFIN.
That is, the intervention constraint module can make distribution adjustments to the data collected by non-random treatment assignment, and the self-interaction module and treatment-aware interaction module can capture different sensitive features of the users in natural and treatment situations, respectively.
\begin{table*}[htbp]
\caption{Ablation study of our EFIN on CRITEO-UPLIFT.}
\centering
\scalebox{1}{
\begin{tabular}{c|cccc}
\specialrule{0.1em}{3pt}{3pt}
 {Dataset} & \multicolumn{4}{c}{CRITEO-UPLIFT}\\
 \specialrule{0.05em}{1pt}{1pt}
 Metrics & LIFT@30 & QINI & AUUC & WAU \\
\specialrule{0.05em}{1pt}{3pt}
w/o Self-Interactive Module & 0.0467 & 0.1266 & 0.0506 & 0.0104\\
w/o Treatment-aware Interaction & \textbf{0.0484} & 0.1254 & 0.0501 & 0.0107\\
w/o Intervention Constraint Module & 0.0442 & 0.1172 & 0.0465 & 0.0109\\
EFIN & 0.0468 & \textbf{0.1285} & \textbf{0.0514} & \textbf{0.0122}\\
\specialrule{0.1em}{1pt}{1pt}
\end{tabular}}
\label{tab:ablation}
\end{table*}


\subsection{RQ3: Results of the Online Deployment}\label{sec:experiments:online}
To further evaluate the performance, we deploy our EFIN on credit card payment scenario in FiT Tencent, which is one of the large-scale online financial platform in China. 
\subsubsection{System Overview \& Scenario Description}
\begin{figure*}[htbp]
    \centering
    \includegraphics[width=.6\linewidth]{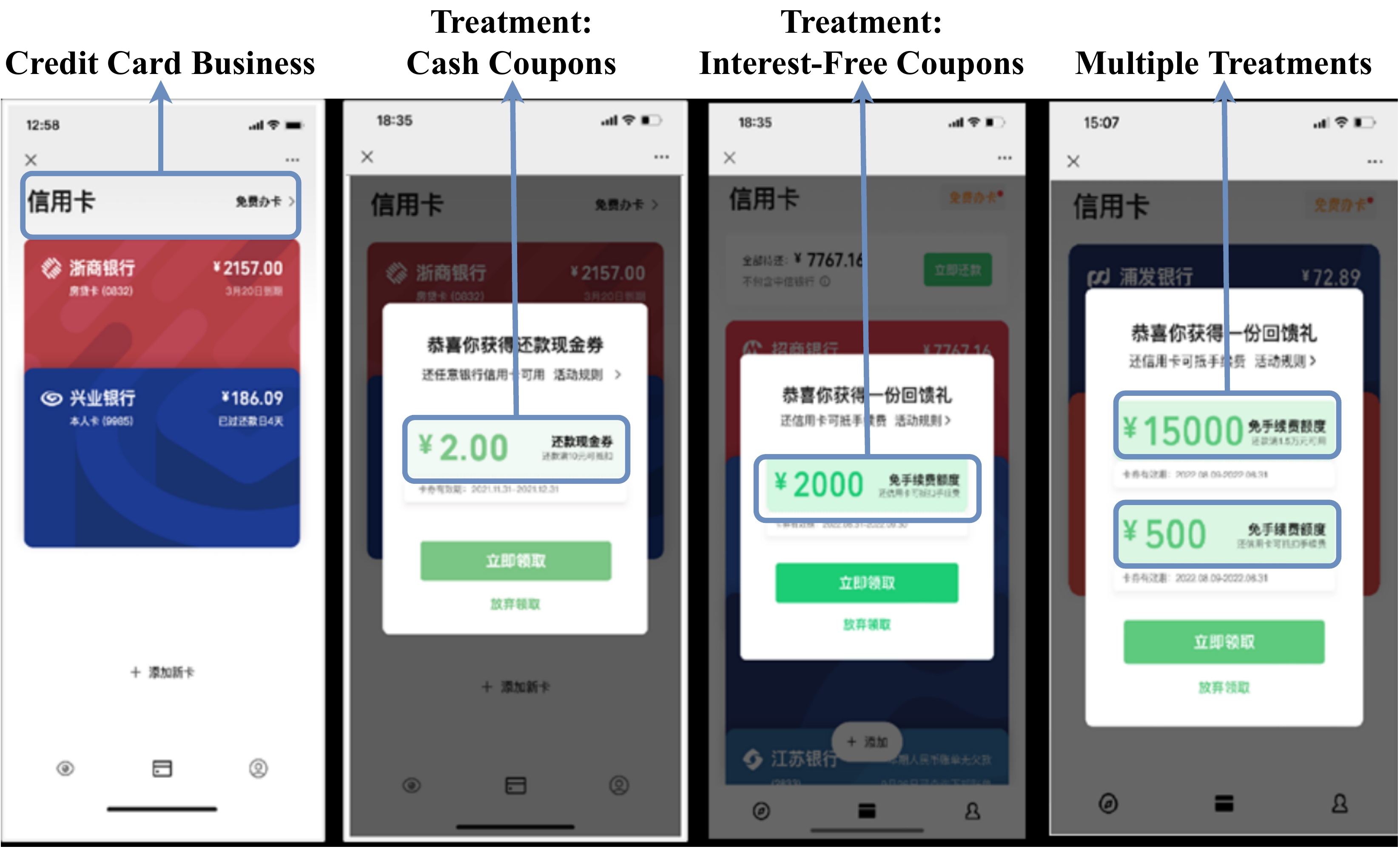}
    \Description[<Figure 4. Fully described in the text.>]{<A full description of Figure 4 can be found in Section 5.4.1 of the text.>}
    \caption{The illustration of credit card scenario.}
    \label{fig:product}
\end{figure*}

\begin{figure}[htbp]
    \centering
    \includegraphics[width=\linewidth]{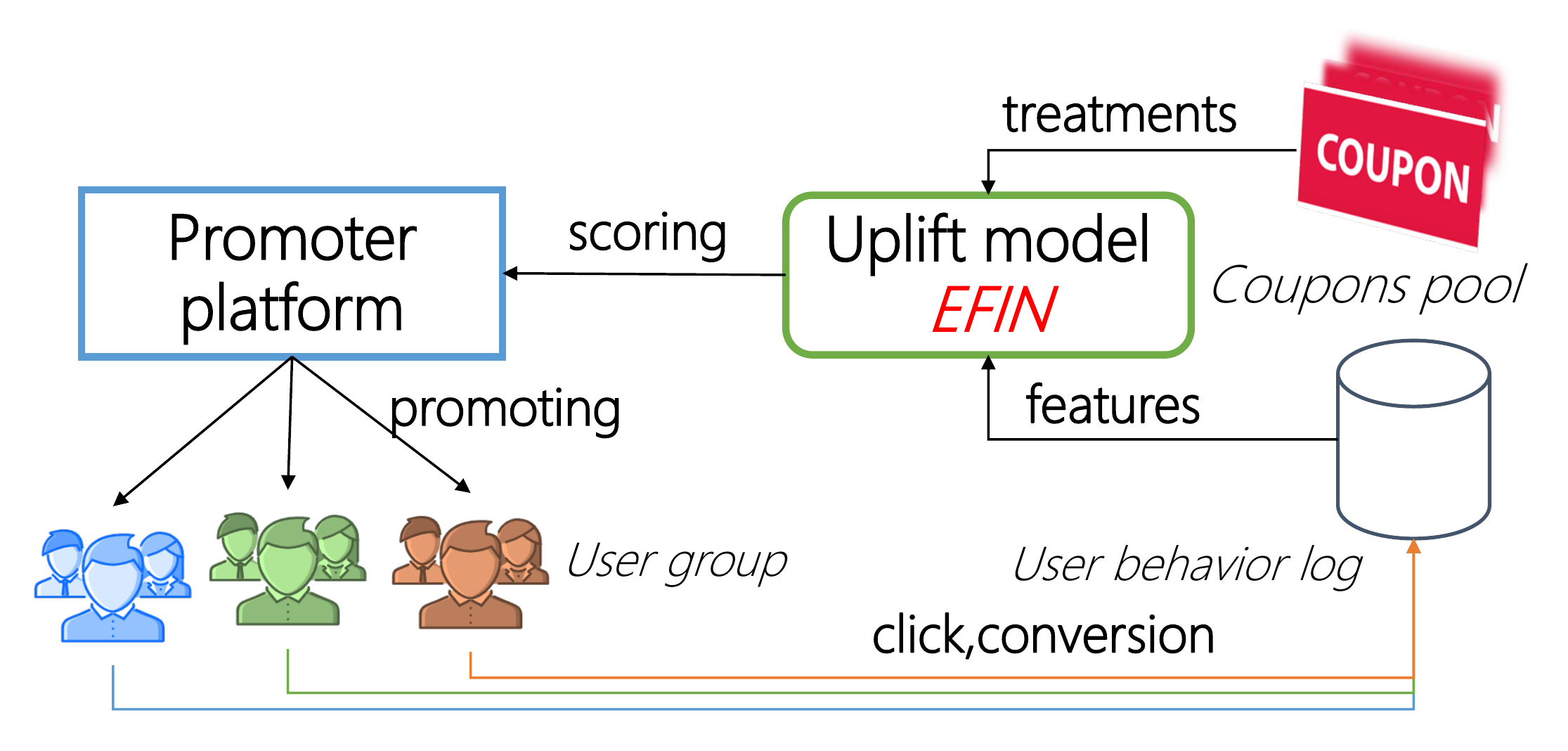}
    \Description[<Figure 5. Fully described in the text.>]{<A full description of Figure 5 can be found in Section 5.4.1 of the text.>}
    \caption{Overview of the promotion system in FiT Tencent.}
    \label{fig:arc}
\end{figure}

The scenario is illustrated in Figure \ref{fig:product}. Marketing in this scenario needs to launch different campaigns for different customer groups to motivate more users to pay the credit card bill within this platform. The treatments are promoted to some group of users once user tends to pay credit card bill on the platform. There are various types of coupons according to the constraints on bill amount in this scenario. Specifically, there are some small denomination coupons without a minimum amount requirement, and some higher denomination coupons requires minimum amount, thus the final number of treatments in this scenario is set as 7. The high-level architecture can be show in Figure \ref{fig:arc}. The user behaviors are pulled from data sources (some storage cluster) to generate features using Apache Spark\footnote{https://spark.apache.org/} and the candidate of treatments is coupons with various amount. The uplift model will score the uplift value for every user on each coupon. Finally, the promoter platform will further deliver coupons to user group with some resource constraints. Notice that our work focus on how to improve the performance of uplift model, which is the key component in the whole system. 
\subsubsection{Online Experimental Results}
To conduct online A/B experiment, we divide two sets of online traffic that do not affect each other, which involved hundreds of millions of users. The existing baseline on the online platform is a multi-head extended T-learner, where each estimator employs XGBoost~\cite{chen2016xgboost} for computation. The baseline model and our EFIN are served for each independent online traffic for one month. As to evaluate the performance, we use two important online metrics: the marketing return on investment (\textbf{ROI}) and the number of monthly active users (\textbf{MAU}). Table \ref{tab:online} reports the relative improvements over the baseline. From the results in Table \ref{tab:online}, we can find that our EFIN improves ROI and MAU by 10\% and 8\% compared to the baseline, respectively.
This means that our EFIN can maintain a stable performance advantage over a period of time, and can indeed accurately capture the sensitive characteristics of different users to perform a reasonable treatment assignment.

\begin{table}[htbp]
\caption{Results of our EFIN in an online deployment.}
\centering
\scalebox{1.1}{
\begin{tabular}{c|cc}
\specialrule{0.1em}{3pt}{3pt}
 Metrics & ROI & MAU \\
\specialrule{0.05em}{3pt}{3pt}
Base (T-Learner + XGBoost) & 0.0\% & 0.0\% \\
EFIN & +10\% & +8\% \\
\specialrule{0.1em}{3pt}{3pt}
\end{tabular}}
\label{tab:online}
\end{table}
\section{Conclusions and Future Work}\label{sec:conclusions}
In this paper, in order to address the underutilization of treatment features and feature interactions that exists in most existing uplift modeling methods, we propose an explicit feature interaction-aware uplift network (EFIN).
Our EFIN consists of four modules, where a feature encoder module is used to encode all features, a self-interaction model aims to accurately model a user's natural response using non-treatment features while isolating treatment information, a treatment-aware interaction module utilizes both treatment features and non-treatment features, and accurately models a user's uplift and response to different treatments through their interactions, and an intervention constraint module is designed to adjust for the distributional differences in the control and treatment groups to make our EFIN more robust across different scenarios.
Finally, we conduct extensive offline and online evaluations and the results validate the effectiveness of our EFIN.

For future work, we plan to explore and analyze the effectiveness of more feature interaction architectures in uplift modeling.
It is also a promising question how to make uplift modeling benefit more from the treatment feature and its interaction with non-treatment features.
In addition, we are also interested in considering and solving some of the more complex uplift modeling scenarios, such as considering necessary constraints like net profit, and modeling a user's response changes to different treatments based on a dynamic perspective.

\begin{acks}
We thank the support of National Natural Science Foundation of China Nos. 61836005, 62272315 and 62172283.
\end{acks}

\bibliographystyle{ACM-Reference-Format}
\bibliography{sample-base}

\appendix









\end{document}